\documentclass[letterpaper, 10 pt, conference]{ieeeconf}  

\IEEEoverridecommandlockouts                              

\usepackage{graphics} 
\usepackage{amsfonts}
\usepackage{siunitx}
\usepackage{xcolor}
\usepackage{gensymb}
\usepackage{graphicx}
\usepackage{amsmath}
\usepackage{amssymb}

\title{\LARGE \bf Vision and Language Navigation in the Real World \\ via Online Visual Language Mapping}

\author{Chengguang Xu, Hieu T. Nguyen, Christopher Amato, Lawson L.S. Wong%
\thanks{Chengguang Xu, Hieu T. Nguyen, Christopher Amato, and Lawson L.S. Wong are with the Khoury College of Computer Sciences,
        Northeastern University, Boston, MA 02115, USA {\tt\small \{xu.cheng, nguyen.trungh, c.amato\}@northeastern.edu; lsw@ccs.neu.edu}, \textit{This work has been submitted to the IEEE for possible publication. Copyright may be transferred without notice, after which this version may no longer be accessible.}}
}

\begin{document}

\maketitle
\thispagestyle{empty}
\pagestyle{empty}

\begin{abstract}
Navigating in unseen environments is crucial for mobile robots. Enhancing them with the ability to follow instructions in natural language will further improve navigation efficiency in unseen cases. However, state-of-the-art (SOTA) vision-and-language navigation (VLN) methods are mainly evaluated in simulation, neglecting the complex and noisy real world. Directly transferring SOTA navigation policies trained in simulation to the real world is challenging due to the visual domain gap and the absence of prior knowledge about unseen environments. In this work, we propose a novel navigation framework to address the VLN task in the real world. Utilizing the powerful foundation models, the proposed framework includes four key components: (1) an LLMs-based instruction parser that converts the language instruction into a sequence of pre-defined macro-action descriptions, (2) an \textit{online} visual-language mapper that builds a real-time visual-language map to maintain a spatial and semantic understanding of the unseen environment, (3) a language indexing-based localizer that grounds each macro-action description into a waypoint location on the map, and (4) a DD-PPO-based local controller that predicts the action.
We evaluate the proposed pipeline on an Interbotix LoCoBot WX250 in an unseen lab environment. \textit{Without any fine-tuning}, our pipeline significantly outperforms the SOTA VLN baseline in the real world.
\end{abstract}

\section{INTRODUCTION}

Humans navigate efficiently in familiar environments by constructing maps containing both spatial and visual contexts (e.g., landmarks) \cite{chun1998contextual, epstein2017cognitive}. For example, humans can easily imagine the path to the coffee machine from anywhere in their houses because they maintain not only a spatial but also a semantic understanding of the environment \cite{epstein2017cognitive}. However, in unfamiliar environments, instructions become necessary for efficient navigation. Therefore, enhancing mobile robots with the ability to follow instructions in natural language will improve navigation efficiently in unseen scenarios, making robots more useful in daily life.

The vision-and-language navigation (VLN) task \cite{anderson2018vision} aims to benchmark this challenge. Specifically, a mobile robot uses visual inputs (e.g., RGB or RGB-D) to navigate in unseen environments by following unstructured natural language instructions. In the initial VLN task, the mobile robot teleports between the nodes on a pre-collected navigation graph of the environment. However, this setting is impractical for real-world robot applications. To address this limitation, \cite{krantz_vlnce_2020, ku2020room} further extends the VLN to continuous environments (VLN-CE) where the robot moves continuously in physical space (i.e. SE(2)) by either taking primitive discrete actions \cite{krantz_vlnce_2020} or by controlling the linear and angular velocities \cite{irshad2021hierarchical}. Despite significant progress being made in VLN-CE, most recent methods \cite{georgakis2022cross, hong2022bridging, chen2022weakly, irshad2022semantically, krantz2021waypoint, krantz_vlnce_2020} are primarily evaluated in simulation, ignoring the complex and noisy real world. 

Transferring a VLN agent trained in simulation to the real world is challenging due to the visual domain gap and the absence of prior environment information \cite{anderson2021sim}. To mitigate these challenges, fusing extra sensor information (e.g. laser scan) and employing domain randomization techniques \cite{tobin2017domain} are recommended \cite{anderson2021sim}. Besides, recent work \cite{huang2023visual, shah2023lm} demonstrates that using foundation models \cite{zhou2023comprehensive}, such as large language models (LLMs) and large visual language models (VLMs), can be beneficial for navigation in the real world. Specifically, LLMs are utilized to parse the instruction into landmarks or executable code, leveraging their powerful textual interpretation capabilities. VLMs are used for processing complex real-world observations and grounding language instructions. 
 However, these methods still require prior mapping of the environment, which is not directly applicable to VLN-CE. 

\begin{figure*}[t!]
    \centering
    \vspace{0.05in}
    \includegraphics[width=.9\linewidth]{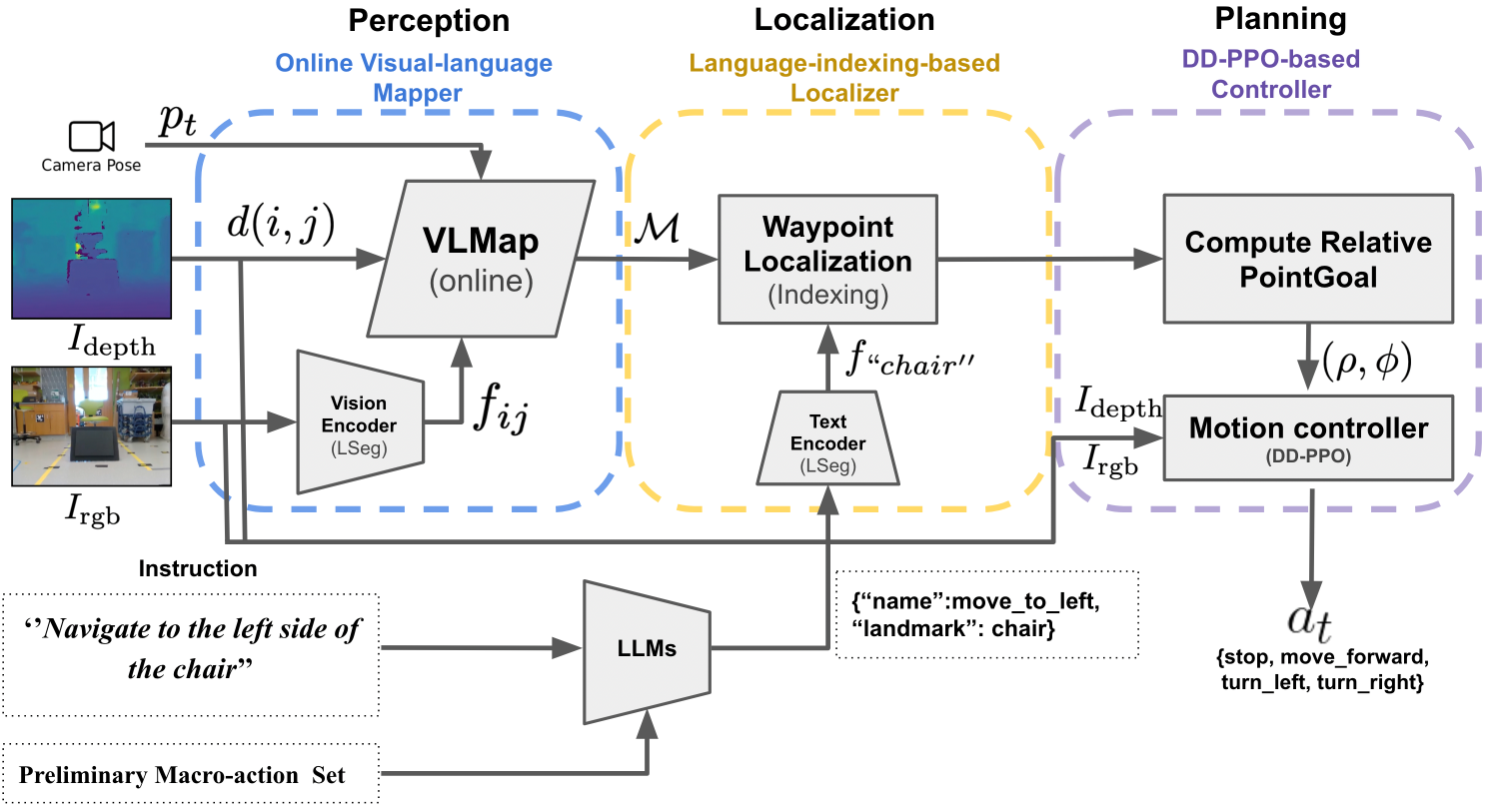}
    \caption{Pipeline overview. Given an instruction in natural language, we first use a large language model (i.e., ChatGPT) to parse it into a sequence of preliminary macro-action descriptions containing both the macro-action name and the related landmark. At every time step, the online visual-language mapper maintains a visual-language map from the front-view RGB-D observations. Using the latest map and the macro-action description, the language-indexing-based localizer outputs the waypoint location on the map. The DD-PPO takes in both the RGB-D observation and a relative point goal, computed from the waypoint location and the agent location on the map, and predicts the next action.}
    \label{fig:framework_overview}
\end{figure*}

In this work, we propose a novel navigation framework to tackle the VLN-CE task in the real world by leveraging powerful foundation models. As depicted in Figure \ref{fig:framework_overview}, to ground the unstructured language instructions, we utilize a large language model (LLM) to parse the instruction into a sequence of pre-defined robot macro-action descriptions, which describe the robot's executable movements and associated landmarks. To handle the complex and noisy observations in unseen environments, we build an \textit{online} visual-language map using a large visual-language model (VLM). With the latest map and the parsed macro-action descriptions, a language indexing-based localizer grounds each macro-action description to a waypoint location on the map. Treating the waypoint as a point goal, we adopt an off-the-shelf DD-PPO local policy to predict the next action. We conducted the experiments on an Interbotix LoCoBot WX250 in an unseen lab environment. In the examined instruction following tasks, \textit{without any fine-tuning}, the proposed pipeline significantly outperforms the state-of-the-art VLN-CE baseline.

\section{RELATED WORK}

\paragraph{Vision-and-Language Navigation}
In VLN, two main settings exist, namely discrete environments (VLN-DE)~\cite{anderson2018vision} and continuous environments (VLN-CE)~\cite{krantz2020beyond}. In VLN-DE, due to the short horizon of an episode, the agent can store the visual memory for every step and efficiently reason the visual memory with language instruction using the attention mechanism~\cite{chen2021history,moudgil2021soat,chen2022think,hong2021vln,qi2021road}. In contrast, the long horizon of an episode in VLN-CE makes the metric map a more reasonable choice of visual memory where the observations at different steps can be fused together in the form of a map \cite{chen2022weakly, georgakis2022cross, irshad2022semantically}. \cite{anderson2021sim} first attempts to tackle the vision-and-language navigation in the real world by transferring the policy trained in the simulator to the real world. Unlike all these works that require training in the simulator, our approach requires \textit{no training} in simulation and \textit{no fine-tuning} in the real world. Instead, we use foundation models trained on large-scale datasets to enable generalization to the real world.

\paragraph{Navigation with online mapping} 
Building maps during navigation achieves impressive performance in multiple embodied navigation tasks such as point-goal navigation, object-goal navigation, and image-goal navigation \cite{chaplot2020learning, karkus2021differentiable, chaplot2020semantic, wu2022image}. However, these methods are designed for particular tasks that require the goal to be specified in a particular format (e.g. a pose, an object class label, or an image). In VLN-CE, only instruction in natural language is provided. Our approach reasons about the goal from the instruction. Besides, the map built by our approach is a visual-language map. Unlike the occupancy maps that store the binary occupancy values and semantic maps that store object class labels, the visual-language map built by our approach stores both the spatial occupancy and the language-related visual features.

\paragraph{Navigation using foundation models} 
The pre-trained foundation models \cite{zhou2023comprehensive} have recently been used in navigation tasks. CoWs \cite{gadre2023cows} adapts open-vocabulary models such as CLIP \cite{radford2021learning} and proposes a language-driven zero-shot object navigation (L-ZSON) task to benchmark object searching. LM-Nav \cite{shah2023lm} proposes a navigation framework that incorporates three types of foundation models (i.e., large language model (LLM), visual-language model (VLM), and visual navigation model (VNM)) and achieves long-distance navigation in outdoor environments. Our approach also uses off-the-shelf foundation models so that \textit{no fine-tuning} is needed during inference. However, our approach is designed for the VLN-CE task, which is different from the L-ZSON task. Unlike LM-Nav, no prior data on the environment is collected for our approach. VLMaps \cite{huang2023visual} is the most related work. But, unlike VLMaps which requires a pre-collected offline dataset to build the environment map beforehand, our method performs online visual-language mapping during navigation. Moreover, both LM-Nav and VLMaps are designed for multi-goal navigation tasks. In summary, our approach can be considered as an extension of VLMaps to tackle the VLN-CE task in the real world.
 
\section{PROBLEM STATEMENT}
We consider the vision-and-language navigation task in continuous environment (VLN-CE) \cite{krantz2020beyond}. In particular, the continuous setting refers to the scenario where the robot has to take primitive actions (e.g., move\_forward, turn\_left) to navigate to the desired goal in physical space (i.e., SE(2)) while following an instruction in natural language. This is in contrast to the discrete setting, where the robot selects discrete nodes from a pre-collected navigation graph, as seen in previous work  \cite{anderson2021sim, hong2022bridging, krantz2020beyond}. 

Formally, at the beginning of each episode, an instruction in natural language $\mathcal{L} = \langle w_0, w_1, w_2, ..., w_L  \rangle$ is given, where $w_i$ is the token for a single word in the instruction. The robot also receives an initial front-view observation $o_0$ determined by the initial pose $s_0 = \langle x_0, y_0, \theta_0 \rangle$, which defines the robot's position and the heading. Following \cite{krantz2020beyond}, at every time step $t$, the robot chooses one action $a_t$ from a set of four discrete actions (i.e., move\_forward, turn\_left, turn\_right, and stop) to execute. Note that, concurrent work like \cite{anderson2018vision} defines the action space as linear and angular velocities, which is different from the current setting. After taking the action $a_t$, the robot moves to a new pose $s_{t+1}$ and observes a new $o_{t+1}$. Following the instruction $\mathcal{L}$, the episode terminates when the robot chooses the ``stop'' action or meets the timeout. The goal is to find a sequence of $\langle s_0, o_0, a_0, s_1, o_1, a_1, ..., s_T, o_t, a_t \rangle$ that aligns with the language instruction $\mathcal{L}$.  

\section{METHOD}

In this section, we first explain how to parse the instruction into macro-action descriptions using LLMs in \textcolor{red}{Sec. \ref{sub_sec:llm_parser}}. Then, the online visual-language mapping is explained in \textcolor{red}{Sec. \ref{sub_sec:visual_language_map}}. Given the latest map and the parsed macro-action descriptions, we explain the language indexing-based localizer in \textcolor{red}{\ref{sub_sec:waypoint_proposer}}. Finally, we discuss the DD-PPO-based local controller in \textcolor{red}{Sec. \ref{sub_sec:ddppo_controller}}.

\subsection{Instruction parser}
\label{sub_sec:llm_parser}
We observe that the instruction in the VLN-CE task consists of several sub-instructions. For instance, in the Room-to-Room (R2R) task \cite{anderson2018vision}, the robot is asked to move from one room to another adjacent room following the instruction. A typical instruction might read as follows ``\textit{Exit the bedroom and turn left. Walk straight passing the gray couch and stop near the rug.}''. The entire instruction can be parsed into a sequence of sub-instructions such as $\langle$\textit{``Exit the bedroom''}, \textit{``Turn left''}, \textit{``Walk straight passing the gray couch''}, \textit{``stop near the rug''} $\rangle$. Furthermore, we have noticed that each parsed sub-instruction describes either a pure robot movement (e.g., ``turn left'') or describes both the movement and associated landmarks. For instance, ``Walk straight passing the gray couch'' contains the movement ``walk straight'' and the landmark ``gray couch''. However, these parsed sub-instructions can not be directly executed by the robot.   

To address this, we leverage the powerful textual interpretation abilities of LLMs (i.e., GPT 3.5 \cite{ouyang2022training}) to parse and convert the instruction into a sequence of pre-defined robot macro-action descriptions. Specifically, inspired by \cite{huang2023visual}, we define a set of macro-action descriptions serving as prior information about the robot's movements.  Formally, we define 10 macro-action descriptions, each represented as a Python dictionary that includes the movement's name and associated parameters. For example, ``Walk straight passing the gray couch'' corresponds to ``\{"name": "move\_to", "landmark": "gray couch"\}''. Following a similar approach to \cite{huang2023visual}, we interact with ChatGPT through few-shot prompt engineering and parse each instruction before conducting navigation experiments.

\subsection{Online visual-language Mapper}
\label{sub_sec:visual_language_map} 
In VLN-CE, collecting data from the target environments is prohibitive because they are assumed to be unseen. Therefore, we extend VLMaps to the online setting and introduce an online mapper that progressively builds the visual-language map of the unseen environment.

In general, the visual-language map fuses the visual-language feature computed from VLMs with a 2-D occupancy grid \cite{huang2023visual}. These visual-language features enhance the representation of the 2-D occupancy map by incorporating richer semantic features compared to semantic labels. Furthermore, the visual-language map inherently benefits from the powerful generalization abilities of VLMs, which are promising to handle complex real-world observations and diverse language instructions. We adopt LSeg \cite{li2022languagedriven, huang2023visual}, a large visual-language model renowned for its dense pixel-wise semantic segmentation driven by flexible language labels. Specifically, LSeg's ViT-based \cite{dosovitskiy2020image} visual encoder aligns the pixel embedding with the text embedding of the corresponding semantic class \cite{li2022languagedriven}. Additionally, LSeg's CLIP-based \cite{radford2021learning} text encoder provides a flexible representation that generalizes well to previously unseen semantic classes during inference. Pretrained on large-scale image-text pairs, LSeg demonstrates significant potential for handling complex robot observations and unseen landmark objects in the real world.

Formally, the visual-language map takes the form of a grid map, denoted as $\mathcal{M} \in \mathbb{R}^{H \times W \times C}$, where $H, W$ are the height and the width of the map, respectively, and $C$ is the dimension of the stored visual-language feature in each grid cell. The resolution of the map is set at $\rho = \SI{5}{\centi\metre}$, and each grid cell corresponds to a \SI{25}{\centi\metre\squared} region in the real world. In contrast to \cite{huang2023visual} that builds the map from an offline dataset, we update the map at every time step. Formally, at the time step $t$, the robot observes a new RGB image $I_{rgb}$, a new depth image $I_{depth}$, and a relative pose change $p_t = \langle x_t, y_t, \theta_t \rangle$ with respective to the initial pose. We assume the knowledge of the camera intrinsic matrix $K$. Consequently, we begin by back-projecting each pixel $(i, j) \in I_{depth}$ into a 3-D point $ p_{cam} = (x, y, z) = K^{-1}(i \times d(i, j), j \times d(i, j), d(i, j))^T$ in the camera frame, where $z = d(i, j)$ is the depth value for pixel $(i, j)$. Then, we project the 3-D points to the world frame  $p_{world} = T^{-1} \times p_{cam}$, where $T$ is the extrinsic matrix. In our world frame definition, the origin is positioned at the top-left corner of the map, the $x$ axis extends to the right, and the $y$ axis extends downward, following the conventions established in  \cite{chaplot2020learning, chaplot2020semantic}. On the map, the robot is consistently initialized in the middle and facing to the right $\langle\frac{H}{2}, \frac{W}{2}, 0.0 \rangle$. Finally, the 3-D points $P_w$ are projected to the map plane as follows:
\begin{equation}
    (p_{map}^x, p_{map}^y) = [\frac{p_{world}^y}{\rho}, \frac{p_{world}^x}{\rho}] 
\end{equation}
Meanwhile, we use the visual encoder of LSeg $E_{\text{ViT}}: \mathbb{R}^{h \times w \times 3} \rightarrow \mathbb{R}^{h \times w \times C}$ to compute the dense pixel-wise visual-language features. Following the same transformation above, we store the pixel-wise visual-language feature $f_{ij} = E_{ViT}(I_{rgb}[i, j])$ of pixel $(i, j)$ at the corresponding grid $(p_{map}^x, p_{map}^y)$. In this way, the new visual-language features are projected onto the map plane as $\overline{M}$. The global map $\mathcal{M}_{t-1}$ gets updated as follows:
\begin{equation}
\mathcal{M}_t[u, v] = 
\begin{cases}
    \overline{M}[u, v], \text{ if } \mathcal{M}_{t-1}[u, v] = \text{None} \\
    \frac{\overline{M}[u, v] + \mathcal{M}_{t-1}[u, v]}{n+1}, \text{otherwise}
\end{cases}
\end{equation}
where $u$, $v$ are grid cell indices and $n$ is the number of stored features. As \cite{huang2023visual}, we average the features in each grid cell to handle the situation where the same object might be perceived from different views.  

\subsection{Language indexing-based localizer}
\label{sub_sec:waypoint_proposer}
The instruction parser parses the instruction into a sequence of macro-action descriptions. Since we use DD-PPO as the local policy, we propose to ground each macro-action description to a waypoint location on the map and set it as an intermediate point goal for DD-PPO. Formally, suppose the current robot location on the map is $\langle x_t, y_t, \theta_t \rangle$.

For pure movement macro-action description such as ``\{"name": "move\_forward", "dist": D\}'', the waypoint position is computed as $\langle x_t + D \times \cos{\theta_t}, y_t + D \times \sin{\theta_t}, \theta_t \rangle$. When there is no specified moving distance, we set the default moving distance to be half meters. A similar strategy is applied to pure turning actions. 

\begin{figure*}[h]
    \centering
    \includegraphics[width=.9\linewidth]{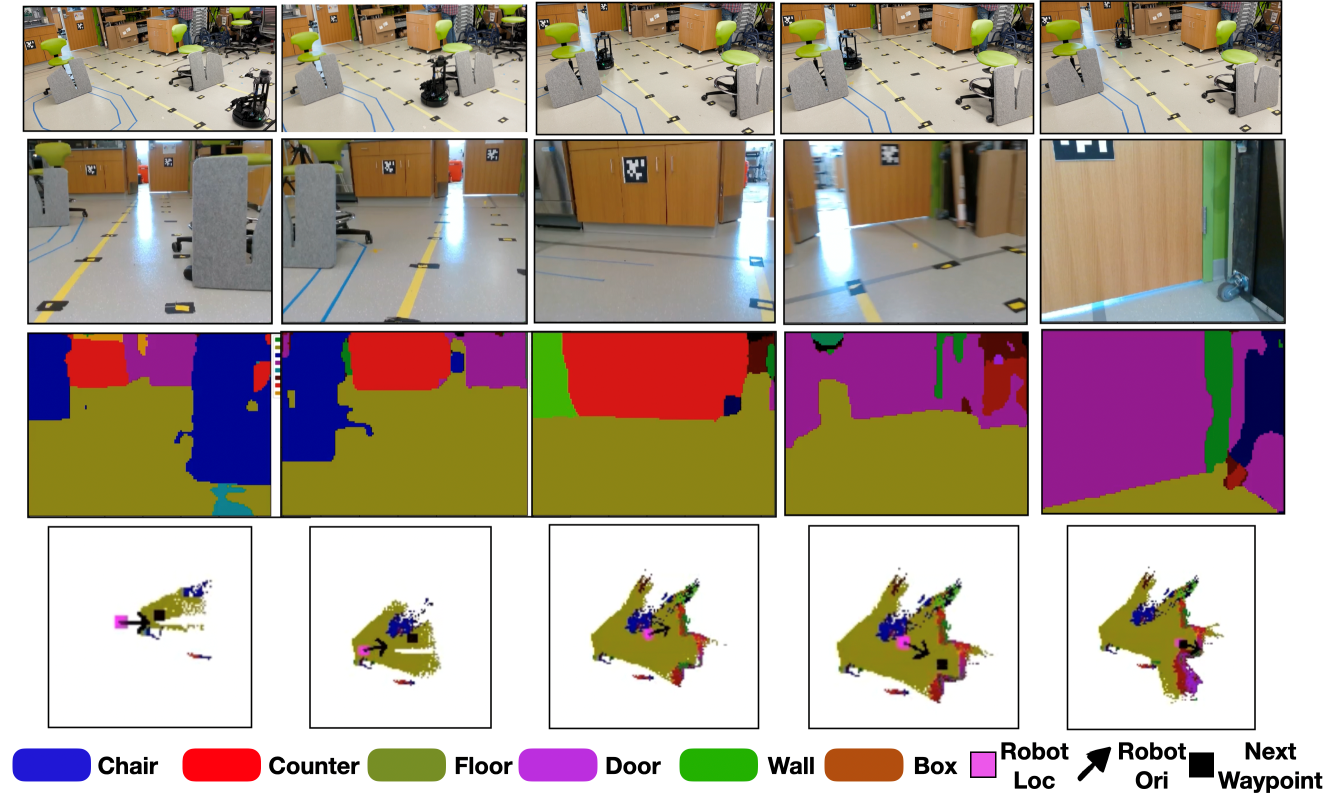}
    \caption{We present a visualization of a successful episode in the complex instruction following task. The first row displays a third-person view showing the robot successfully follows the complex instruction and reaches the final goal. The second row shows the robot's front camera view at every waypoint during the navigation. The third row provides a real-time semantic visualization to demonstrate the effective generalization of VLMs to real-world observations. The final row shows the visual-language map constructed by our online mapper. On the map, the waypoints (black squares) illustrate our method's success in grounding the instruction into spatial point goals.}
    \label{fig:visualization_traj}
\end{figure*}

For landmark-associated macro-action such as ``\{"name": "move\_to\_left", "landmark": "chair"\}'', we first localize the landmark object on the visual-language map through language indexing. Specifically, we construct a label list $[l_{\text{target}}, l_{\text{default}}^2, ..., l_{\text{default}}^k, \text{other}]$ where the first world is the landmark label and the remaining are the default labels. Note that ``other'' is LSeg's default label to represent any out-of-range object classes. LSeg's text encoder takes in the label list and outputs a text embedding feature matrix $f_\text{text} \in \mathbb{R} ^ {C \times (K + 1)}$. The similarity score for every label at every grid cell can be computed as $\mathcal{M}_t \times f_\text{text}$, where $\mathcal{M}_t \in \mathbb{R}^{H \times W \times C}$. With the similarity matrix, we choose the label for each grid cell by selecting the label with the maximal similarity score. Therefore, at every time step, a semantic map is generated. To localize the desired landmark, we first apply density-based spatial clustering (DBSCAN) \cite{ester1996density} to find the centers of all landmark labels. Next, we compute the orientation and Euclidean distance between the robot's current location and the centers on the map. We select the nearest label in front of the robot and use the corresponding center location as the waypoint. The design choice is because the instructions in VLN-CE are generated from the perspective of the robot's egocentric view. Combined with online mapping, we can mitigate the object ambiguity issue during navigation (See Figure \ref{fig:visualization_traj}). The waypoint is represented as a 2-D egocentric polar coordinate $(\rho, \phi)$, where $\rho$ represents the relative distance of the waypoint in meters and $\phi$ is the egocentric orientation towards the waypoint in radius.

\subsection{DD-PPO-based local controller}
\label{sub_sec:ddppo_controller}
To deal with the noisy observations in the real world, we use the DD-PPO navigation policy, pre-trained on a large-scale point-goal navigation task, as the local controller \cite{partsey2022mapping, wijmans2019dd}. Specifically, the controller takes in a front-view RGB-D observation $\{I_{rgb}, I_\text{depth}\}$ and a point-goal represented as a 2-D egocentric polar coordinate $(\rho, \phi)$ as inputs. The off-the-shelf local policy $\pi(a_t|I_{rgb}, I_{depth}, (\rho, \phi))$ predicts the next action $a_t$. Specifically, the action space is discrete and contains four primitive actions including a ``stop'' action to indicate termination or reaching the goal point.

\section{EXPERIMENTS}
\subsection{Mobile robot and environment setup}
\begin{figure*}[h]
    \centering
    \includegraphics[width=.8\linewidth]{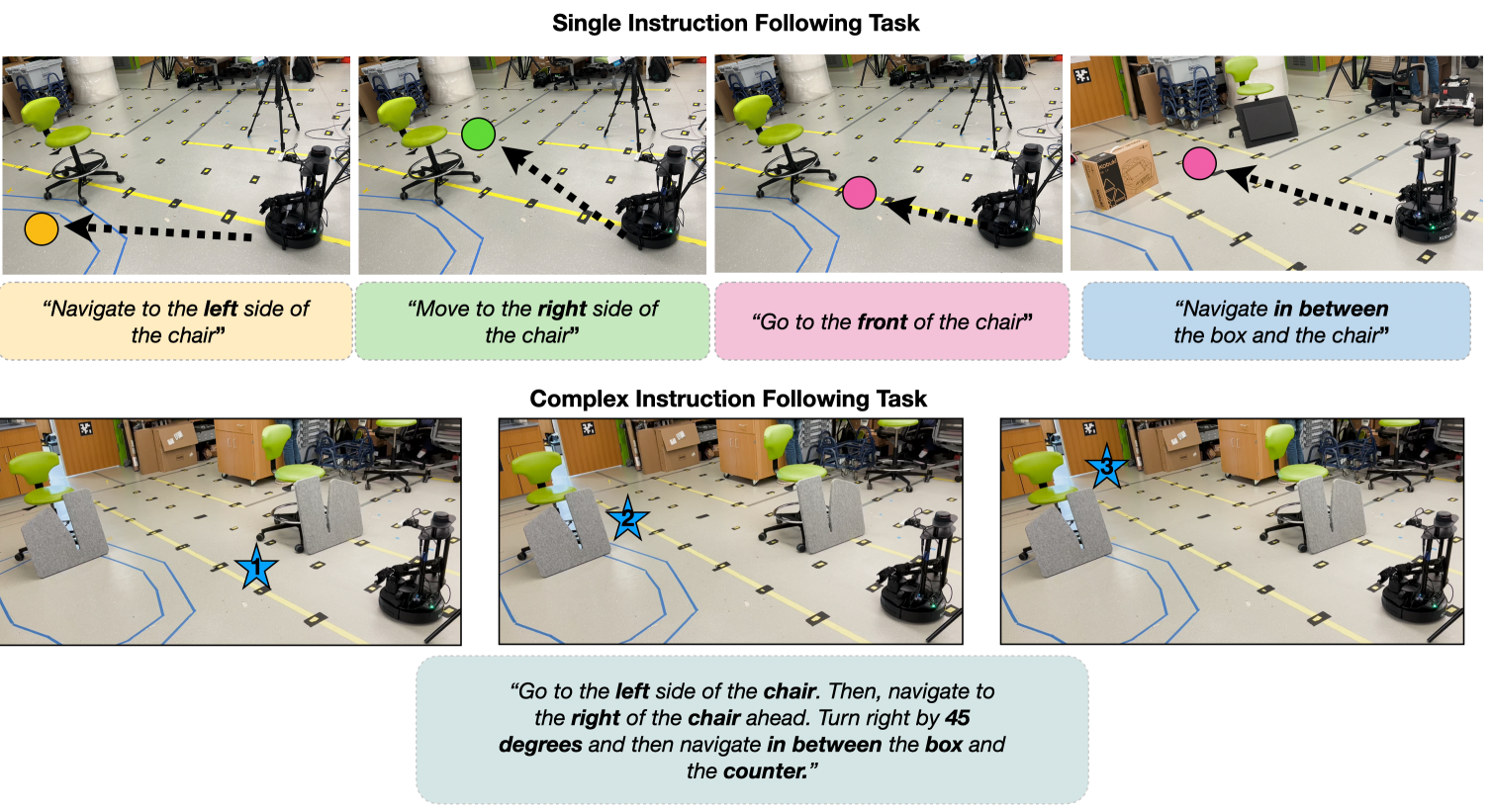}
    \caption{We introduce two types of instruction following tasks. (1) The 'single instruction' task assesses the robot's capability to accurately infer the goal location from a single instruction and execute the task correctly. (2) The complex instruction following task further examines the ability to ground longer instructions in the real world. We provide visualizations of instances for both of these tasks here.}
    \label{fig:task_overview}
\end{figure*}

We conducted all experiments using an Interbotix LoCoBot WX250 equipped with an Intel RealSense D435 camera for capturing both depth and RGB images. The RGB image dimensions are $640 \times 480 \times 3$ and the depth dimensions are $640 \times 480$. The camera is mounted on a Kobuki base at a height of approximately $\SI{53}{\centi\metre}$ with an elevation angle of $-15.7$ degrees. In our experiments, we disabled the robot's arms and exclusively controlled the Kobuki base. We implemented four primitive actions to align with the output of the DD-PPO local policy. Specifically, the 'move\_forward' action advances the robot by $\SI{0.25}{\centi\metre}$, while the 'turn\_left' and 'turn\_right' actions rotate the base by $15$ degrees. A 'stop' action was also included for no movement. We manage the whole experiment on ROS Noetic. We conducted the entire experiment using ROS Noetic in an unstructured lab environment, which was unseen by both our pipeline and CM2. Figure \ref{fig:task_overview} shows the lab environment and the robot.
\subsection{Baseline}
We compare our method against Cross-modal Map Learning (CM2)~\cite{georgakis2022cross}, a learning-based state-of-the-art method to tackle the VLN-CE task. CM2 employs a strategy where it generates both global occupancy maps and global semantic maps by hallucinating information from local maps back-projected from depth and semantic observations, enhancing the spatial and semantic understanding of the unseen environments. Given an instruction, CM2 learns cross-modal map attention to ground the entire instruction into a sequence of waypoints on the map. The waypoint sequence will be predicted at every time step and the same DD-PPO local policy is used to predict the next action. We select the best model provided by the author as our comparison \footnote{https://github.com/ggeorgak11/CM2}. To make the comparison fair, both CM2 and our method use the same front-view RGB-D observations and relative pose as inputs. We also use the same DD-PPO controller from \cite{wijmans2019dd, partsey2022mapping}. It's important to note that CM2 is extensively trained in simulation and does not undergo fine-tuning with real-world data. In contrast, our pipeline requires no training in the simulator and no fine-tuning in the real world, by directly leveraging pre-trained foundation models. 

\subsection{Instruction following tasks} 
In the context of VLN-CE, instructions consist of multiple sub-instructions that the robot must follow. To evaluate the performance of our proposed pipeline, we designed two types of instruction-following tasks, ranging from easy to challenging. The first task, 'single instruction following,' involves instructions that contain only one sub-instruction for the robot to execute. For example, an instruction might read, 'Walk forward by 2 meters.' This task aims to assess the robot's ability to correctly infer the goal from the instruction and execute it accurately. It's important to note that in VLN, the goal location is implicitly encoded in the natural language instruction. The second task, 'complex instruction following,' is more demanding. Here, instructions include multiple sub-instructions that the robot must carry out. For instance, an instruction might involve steps like 'Move to the left side of the chair. Then, turn left by 90 degrees.' In addition to evaluating goal inference and execution accuracy, this task assesses each method's ability to ground complex instructions in the real world.

\subsection{Results}

\paragraph{\textit{Single Instruction Following - Pure Motion Task}}: We evaluate the accuracy of our method in executing instructions that involve pure movement. The tested straight distances range from $0.5$ to $2.0$ meters, with each distance tested in 5 independent runs using different instructions. For example, instructions include ``\textit{Go forward by $1.0$ meter}'' or ``\textit{Navigate ahead by $1.0$ meter.}'' We provide two metrics: 'Actual Dist,' representing the straight distance actually traversed by our pipeline, and 'Est Dist,' an estimate of the distance between the stop position and the goal position on the map. The average movement error is approximately $\SI{1.4}{\centi\metre}$. This result highlights the effectiveness of using DD-PPO as the local policy in real-world scenarios and the accuracy of the map built by our online mapper.

\begin{table}[h]
\begin{center}
\caption{Results of Pure Motion in \textit{Single Instruction Following} Task}
\label{table_1_single_instructin_follow_motion}
\begin{tabular}{|c|c|c|c|}
\hline
Desired Dist (m) &  Actual Dist (m) & Est Dist (m) & Err Dist (m)\\
\hline
0.5 & 0.426 & - & - \\
1.0 & 0.748 & 0.238 & 0.014 \\
2.0 & 1.678 & 0.308 & 0.014 \\
\hline
\end{tabular}
\end{center}
\end{table}

\paragraph{\textit{Single Instruction Following - Landmark-Associated Task}}: The instruction implicitly specifies a spatial goal. We evaluate our method using four instructions, indicating four distinct spatial goals. For each instruction, we conduct 5 independent runs with varying initial robot locations. In Table \ref{table_2_single_instruction_follow_landmark}, our approach significantly outperforms the CM2 baseline, achieving a much higher mean success rate of $95\%$ compared to CM2's $30\%$ and substantially smaller distances to the goal location (Ours $\SI{0.2}{\meter}$ v.s. CM2's $\SI{2.06}{\meter}$). The key to our method's success lies in the powerful generalization ability of VLMs to real-world observations. In contrast, CM2 struggles with generalization to the real world due to the visual domain gap, despite it using the same DD-PPO local controller.

\begin{table}[h]
\begin{center}
\caption{Results of Landmark-associated Motion in \textit{Single Instruction Following} Task}
\label{table_2_single_instruction_follow_landmark}
\begin{tabular}{|c|c|c|c|}
\hline
Method & \multicolumn{2}{c|}{CM2 \cite{georgakis2022cross}} \\
\hline
Instruction &  SR ($\%$) & Dist to Goal (m)\\
\hline
``Navigate to the \textit{left} side of the chair''  & 60 & 0.88 \\
``Navigate to the \textit{right} side of the chair'' & 40 & 0.97 \\
``Navigate to the \textit{front} of the chair''      & 20 & 1.37 \\
``Move \textit{in between} the box and the chair''   &  0 & 2.06 \\
\hline
Average & 30 & 1.32 \\
\hline
Method & \multicolumn{2}{c|}{Ours} \\
\hline
``Navigate to the \textit{left} side of the chair''  & 100 & 0.79 \\
``Navigate to the \textit{right} side of the chair'' & 100 & 0.83 \\
``Navigate to the \textit{front} of the chair''      & 100 & 0.81 \\
``Move \textit{in between} the box and the chair''   &  80 & 0.20 \\
\hline 
Average & \textbf{95} & \textbf{0.66} \\
\hline
\end{tabular}
\end{center}
\end{table}

\paragraph{Complex Instruction Following}: We presented the robot with a complex instruction comprising multiple sub-instructions and landmark objects. The instruction used was: ``\textit{Go to the left side of the green chair. Then, navigate to the right of the red chair ahead. Turn right by 45 degrees and then navigate in between the box and the counter.}'' We conducted the experiment with 5 independent runs from different initial robot locations. Table \ref{table_3_complext_instruction} demonstrates the results of this task. CM2 struggled to complete this complex instruction, likely due to the visual domain gap making waypoint prediction challenging. In contrast, our method achieved a 100\% success rate and stopped approximately $\SI{26}{\centi\meter}$ from the goal location, despite using the same DD-PPO controller as CM2.

\begin{table}[h]
\caption{Results in \textit{Complex Instruction Following} Task}
\label{table_3_complext_instruction}
\begin{center}
\begin{tabular}{|c|c|c|c|}
\hline
Method &  SR ($\%$) & Dist to Goal (m) & Time steps\\
\hline 
CM2 \cite{georgakis2022cross} & 0 & 4.9 & 203.4\\
\hline
Ours &\textbf{ 100} & \textbf{0.256} & \textbf{88.6} \\ 
\hline
\end{tabular}
\end{center}
\end{table}

In summary, CM2 achieves the state-of-the-art performance of VLN-CE in the simulation but shows limited generalization to the VLN-CE task in the real world. Using foundation models (LLMs, VLMs, and DD-PPO), our method achieves much better performance than transferring a VLN-CE navigation policy to the real world.

\section{CONCLUSIONS}

We propose a novel navigation framework to handle the VLN-CE task in the real world via online visual-language mapping. Using three foundation models (LLMs, VLMs, and DD-PPO), \textit{with no training required}, our method significantly outperforms the state-of-the-art VLN-CE baseline in the real world. We argue that the visual domain gap between simulation and the real world poses challenges for language grounding and sim-to-real transfer, especially when instructions are complex. We hope that our pipeline can shed some light on using large foundation models to resolve the VLN-CE task in the real world. While our pipeline offers promising results, it has limitations. First, when the instruction becomes more complex, the grounding performance might be limited by the scope of the pre-defined macro-action descriptions. Second, the localizer in our method prefers landmark objects or absolute movement for better waypoint proposals. However, instructions can be coarse and high-level. 

\clearpage

\addtolength{\textheight}{-12cm}   







\bibliographystyle{IEEEtran}
\bibliography{IEEEexample}

\clearpage

\end{document}